\documentclass[9pt,twocolumn]{article}
\usepackage{./spconf}
\usepackage{amsmath,amssymb}
 \usepackage{mathrsfs}
\usepackage{dsfont}
\usepackage{tabularx}
\usepackage{graphicx}
\usepackage{color}

\usepackage[font=scriptsize,labelfont=bf]{caption}

\usepackage{./Tianpei_Report}
\usepackage{appendix}
\usepackage{algorithm}
\usepackage{algorithmic}

\DeclareMathOperator*{\minimize}{\text{minimize}}

\DeclareMathOperator*{\st}{\text{subject to}}

\begin{document}
\allowdisplaybreaks
\setlength{\textfloatsep}{5pt}
\setlength{\abovedisplayskip}{2pt}
\setlength{\belowdisplayskip}{3pt}
\title{Semiblind subgraph reconstruction in Gaussian graphical models}
\name{Tianpei Xie, \hspace*{0.1in}Sijia Liu\hspace*{0.1in}  and \hspace*{0.085in} Alfred O. Hero III \thanks{Acknowledgement: This research was partially supported by US Army Research Office (ARO) grants W911NF-15-1-0479 and W911NF-15-1-0241.}}
\address{Dept. of Electrical Eng., University of Michigan, Ann Arbor, MI 48109 \\
$\{$tianpei,\; lsjxjtu, \;hero$\}$@umich.edu}
\maketitle
\begin{abstract}
{Consider a social network where only a few nodes (agents) have meaningful interactions in the sense that the conditional dependency graph over node attribute variables (behaviors) is sparse. A company that can only observe the interactions between its own customers will generally not be able to accurately estimate its customers' dependency  subgraph: it is blinded to any external interactions of its customers and this blindness creates false edges in its subgraph.  In this paper we address the semiblind scenario where the company has access to a noisy summary of  the complementary subgraph connecting external agents, e.g., provided by a consolidator. The proposed framework applies to other applications as well, including field estimation from a network of awake and sleeping sensors and privacy-constrained information sharing over social subnetworks.    We propose a penalized likelihood approach in the context of a graph signal obeying a Gaussian graphical models (GGM). We  use a convex-concave iterative optimization algorithm to maximize the penalized likelihood.} 
The effectiveness of our approach is demonstrated through numerical experiments and comparison with state-of-the-art GGM and latent-variable (LV-GGM) methods.  
\end{abstract}

\vspace{-5pt}
\section{Introduction}\vspace{-5pt}
\allowdisplaybreaks
Learning a dependency graph given relational data  is an important task for sensor network analysis \cite{joshi2009sensor, liu2016sensor} and social network analysis \cite{scott2012social}. In many situations, however, a learner may only have access to {data on a subgraph: the learner is blinded to the rest of the graph, e.g.,} due to energy constraints or privacy  concerns. For instance,  in a sensor network with limited power budget \cite{joshi2009sensor, liu2016sensor}, a subset of sensors that were actively collecting data  in the recent past may have gone into sleeping mode at the current time. As a result, the fusion center   only acquires measurements from the active sensors  at the current time. 
In this scenario, without any information regarding the {unobserved external} data, confounding marginal correlations may exist between observed variables that are conditionally uncorrelated. In this paper, we consider {the semiblind scenario} where, in addition to  the observed internal data (e.g., measurements of active sensors),  the learner  receives a noisy summary about   partial correlations  of  data from {unobserved external nodes} (e.g., previously measured spatial correlation between sleeping sensors). We call this the semiblind scenario. The goal of this paper is to learn the network topology using the observed internal data as well as the partially shared information from external sources.  

We consider a random graph signal that follows Gaussian graphical model (GGM)  \cite{koller2009probabilistic}. The GGM can be learned efficiently via  sparse inverse covariance  estimation  \cite{meinshausen2006high,  friedman2008sparse, hsieh2011sparse, marjanovic2015l0}. However, due to  effects of marginalization \cite{koller2009probabilistic} over latent factors {(the unobserved external nodes)}, these aforementioned methods suffer from a significant loss in terms of estimation accuracy. In \cite{chandrasekaran2012latent}, a latent variable Gaussian graphical model (LV-GGM) was proposed to learn the sparse target sub-network  via sparse and low-rank matrix separation.  Theoretical analysis  \cite{chandrasekaran2012latent, meng2014learning} shows that the maximum likelihood estimator of LV-GGM is unbiased under some mild conditions. 
{The LV-GGM proposed in \cite{chandrasekaran2012latent, meng2014learning} assumes a blind scenario where no information about the latent variable subgraph is available. Here we treat the semiblind scenario where the learner has access to noisy information about this subgraph  and its associated dependency matrix.   }

Specifically, we propose a  \emph{Decayed-influence Latent variable Gaussian Graphical Model (\textbf{DiLat-GGM})} that in this model 
the influence of each latent variable decays over the underlying network. Examples include, but are not limited to, spatial correlation of sensor networks \cite{nowak2004estimating} and propagation of social influence \cite{goyal2010learning}. In contrast with LV-GGM, DiLat-GGM  {leads to a special non-convex optimization problem, which is identified with  a difference of convex (DC) program. }An efficient algorithm based on convex-concave procedure (CCP) \cite{lipp2016variations,yuille2002concave} and an {alternating direction method of multipliers (ADMM) \cite{boyd2011distributed}} is employed to find a locally optimal solution. Extensive experiments  are provided to show the superiority of DiLat-GGM over existing methods in terms of estimation  accuracy  in the semiblind scenario. 

\vspace{-5pt}



\setlength{\abovedisplayskip}{5pt}
\setlength{\belowdisplayskip}{2pt}
\setlength{\abovedisplayshortskip}{0pt}
\setlength{\belowdisplayshortskip}{0pt}
\section{Preliminaries and Problem formulation}\label{sec: problem_formulation_chap3}\vspace{-5pt}
We begin by introducing some notations. Let $\cG = (\cV, \cE)$ be  an undirected unweighted graph, where $\cV$ is the vertex set with cardinally $|\cV| = n$, and $\cE$ is the edge set. 
Let $\mb{x}=[x_1,\ldots, x_n]$ be a random vector over  vertices in $\cV$,
following the multivariate Gaussian distribution $\mb{x}\sim \cN(\mb{0}, \mb{\Theta}^{-1})$, where $\mb{\Theta}:= \mb{\Sigma}^{-1} \in \bR^{n\times n}$ is the precision matrix, namely, the inverse of covariance matrix $\mb{\Sigma}$. Assume that
  $\mb{x}$ obeys the Markov property with respect to $\cG$, i.e. 
  $\Theta_{ij} = 0$ for all $i \neq j$ and $(i,j) \notin \cE$, where $\Theta_{ij}$ denotes the $(i,j)$-th entry of $\mb{\Theta}$. 

\vspace{-15pt}

\begin{figure}[tb]
  \centering
    \begin{minipage}[b]{1\linewidth}
  \centering
  \centerline{\includegraphics[scale = 0.19]{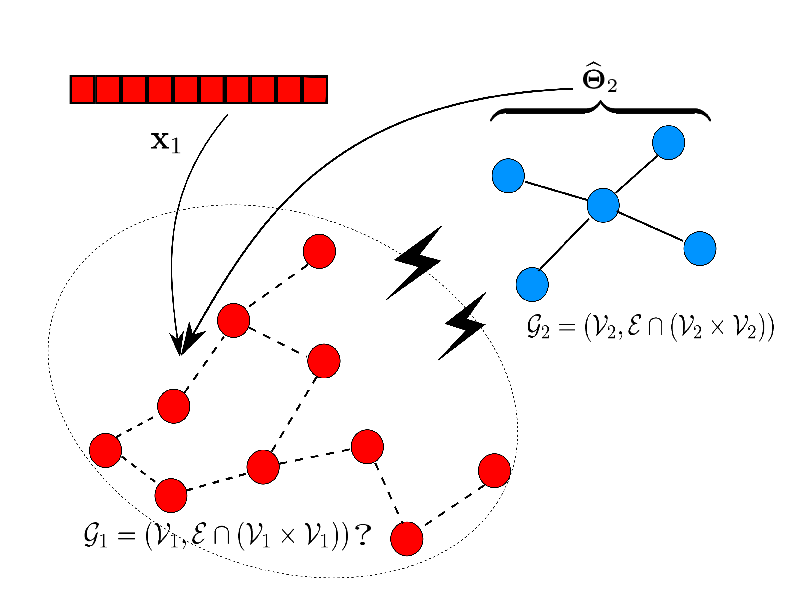}}
  \vspace{-15pt}
  \end{minipage}
  \caption{\scriptsize  The red vertices are affected by the blue vertices through some unknown links.  Data on the blue vertices are not observed directly but a noisy summary $\widehat{\mb{\Theta}}_2$ of the precision matrix whose zero entries spectify the subgraph $\cG_2$.  The task is to infer the subgraph $\cG_1$ from obervations of its node attribute vector $\mb{x}_1$ and  the noisy summary  $\widehat{\mb{\Theta}}_2$.  } \label{fig: demo}
\end{figure}

\subsection{Problem Statement}\label{sec: notations}\vspace{-10pt}
Without loss of generality, the vertex set of $\mathcal G$ is partitioned into two nonoverlapping  subsets $\cV_1$ and $\cV_2$ with   $|\cV_1|=n_1$ and $|\cV_2|=n_2$, respectively. {Here the sub-graph $\cG_1= (\cV_1, \cE\cap (\cV_1 \times \cV_1) )$ is called the \emph{target} sub-network associated with the precision matrix $\mb{\Theta}_1$, and  $\cG_2 =(\cV_2,  \cE\cap (\cV_2 \times \cV_2))$ is called the \emph{external} network with $\mb{\Theta}_2$, where $\mb{\Theta}_{i} \in \mathbb R^{n_i \times n_i}$ denotes the submatrix of $ \mb{\Theta} $ indexed by $\mathcal V_i$ for $i = 1$ and $2$. We assume that only an internal dataset $\mb{X}_1 \in \bR^{n_1 \times m}$ (namely, $m$ i.i.d   samples of $\mb{x}_{\mathcal V_1}$)  and a summary information $\widehat{\mb{\Theta}}_2$ (namely, an estimate of $\mb{\Theta}_{2}$) are available to the learner. Our task is to estimate the network topology of $\cG_1$ (in terms of $\mb{\Theta}_1$) only based on data  $\mb{X}_1$ and the shared summary information $\widehat{\mb{\Theta}}_2$ from the external source $\cG_2$.}  We emphasize that  data samples of $\mb{x}_{\cV_2}$ (latent variables) over $\cG_2$ are not shared with $\cG_1$ due to the energy or privacy concerns.
Figure \ref{fig: demo} provides an overview of the studied problem in this paper.    \vspace{-15pt}

\subsection{Blind Sub-network Inference via LV-GGM}\label{sec: lvggm} \vspace{-8pt}
{In the blind scenario, the learner only has access to $\mb{X}_1$, from which the }marginal covariance matrix $\widehat{\mb{\Sigma}}_1 := \mb{X}_1\mb{X}_1^{T}/m $ {can be constructed.} To recover the sub-network $\cG_1$,   
the following partitioned matrix inverse identity {specifies the relation between the inverse of the mean of $\widehat{\mb{\Sigma}}_1$ (the ensemble marginal covariance) and the associated block $\mb{\Theta}_1$ of $\mb{\Theta}$} \cite{meng2014learning}
\begin{normalsize}
 \setlength{\abovedisplayskip}{5pt}
\setlength{\belowdisplayskip}{2pt}
\setlength{\abovedisplayshortskip}{0pt}
\setlength{\belowdisplayshortskip}{0pt}
\begin{align}
 \widetilde{\mb{\Theta}}_{1}:=\paren{\mb{\Sigma}_{1}}^{-1} &= \mb{\Theta}_{1} - \mb{\Theta}_{12}\paren{\mb{\Theta}_{2}}^{-1}\mb{\Theta}_{21}:=\mb{C} - \mb{M}  \label{eqn: low_rank_sparse_decomposition} 
\end{align}\end{normalsize}where $ \small \widetilde{\mb{\Theta}}_{1}$ is the marginal precision matrix over $\mb{x}_1$, recalling that $\small \mb{\Theta}_{1}$ and $\small \mb{\Theta}_{2}$ are submatrices of the global precision matrix $\small \mb{\Theta}$ over $\cV_1$ and $\cV_2$,
and $\small \mb{\Theta}_{21} \in \mathbb R^{n_2\times n_1}$ is the partial cross-covariance matrix between $\mathcal V_1$ and  $\mathcal V_2$. 
As noted in \cite{chandrasekaran2012latent, meng2014learning}, the marginal precision matrix $ \widetilde{\mb{\Theta}}_{1} $ can be decomposed into a sparse matrix  $ \mb{\Theta}_{1}$, which is associated with $\cG_1$, plus a low-rank matrix that characterizes the effect of marginalization. 

In \cite{chandrasekaran2012latent},  the latent variable Gaussian graphical model (LV-GGM) was introduced to find the separation $( \widehat{\mb{C}}, \widehat{\mb{M}})$  in \eqref{eqn: low_rank_sparse_decomposition}  by maximizing the regularized marginal log-likelihood objective function
 \begin{normalsize}
 \setlength{\abovedisplayskip}{5pt}
\setlength{\belowdisplayskip}{2pt}
\setlength{\abovedisplayshortskip}{0pt}
\setlength{\belowdisplayshortskip}{0pt}
\begin{align}
\hspace*{-0.1in} \begin{array}{ll}
  \displaystyle \minimize_{\mb{C}, \mb{M}}  & -\log\det\paren{\mb{C} - \mb{M}} + \text{tr}\paren{\widehat{\mb{\Sigma}}_{1}\paren{ \mb{C} - \mb{M}}}  \\
  & + \alpha \norm{\mb{C}}{1} + \beta\norm{\mb{M}}{*} \vspace*{0.05in} \\
    \st &  \mb{C} - \mb{M} \succeq \mb{0}, \;\; \mb{M} \succeq \mb{0},
\end{array}
\label{eq: LV}
\end{align}
 \end{normalsize}
\hspace*{-0.052in}where {$\norm{\mb{M}}{*}$ is the nuclear norm of $\mb{M}$, $\tr{\cdot}$ and $\log\det(\cdot)$ are trace and log-determinant operator, $\mb{A}\succeq \mb{0}$ means that $\mb{A}$ is positive semidefinite},  $\small \widehat{\mb{\Sigma}}_1$ is the sample  covariance, $\alpha$ and $\beta$ are regularization parameters for the $\ell_{1}$-norm and the nuclear-norm, respectively. Asymptotic analysis  \cite{chandrasekaran2012latent, meng2014learning} shows that the resulting estimator of LV-GGM is unbiased under some mild conditions.

%

\vspace{-13pt}



\subsection{Semiblind Sub-network Inference  under Decayed Influence}\label{sec: decay_influence}\vspace{-8pt}
In this semiblind scenario the learner has access to both the marginal sample covariance $\widehat{\mb{\Sigma}}_1$ and a noisy version of $\mb{\Theta}_2$. {It is notable that $\mb{\Theta}_2$ can be determined without knowledge of the full covariance matrix $\mb{\Sigma}$: all that is required is the covariance of the buffered external nodes \cite{meng2013distributed}, defined by first-order neighboring nodes of vertices in $\mathcal V_2$.}  Compared with LV-GGM, we consider a decayed influence model: the influence of each node decays while propagating along the graph. {Therefore, latent variables that are at large hop distance from the target network $\cG_1$ have little impact on the inference of $\cG_1$.} This implies that $\mb{\Theta}_{21}$ in \eqref{eqn: low_rank_sparse_decomposition}  maintains row-sparsity structure. Specifically, let $\mb{B} := \mb{\Theta}_{12}\mb{\Theta}_2^{-1} \in \bR^{n_1 \times n_2}$ and $\mb{\mu}_{2|1}:= \mb{B}^{T}\mb{x}_1$ be the conditional mean of $\mb{x}_{\cV_2}$ given $\mb{x}_{\cV_1}$. The  low rank term in \eqref{eqn: low_rank_sparse_decomposition} can be reparameterized as $\mb{M}:= \mb{B}\mb{\Theta}_{2}\mb{B}^{T}$. Motivated by \eqref{eq: LV},
{\textbf{D}ecayed-\textbf{i}nfluence \textbf{Lat}ent variable \textbf{G}aussian \textbf{G}raphical \textbf{M}odel (\textbf{DiLat-GGM})}
is formulated as
\begin{normalsize}
 \setlength{\abovedisplayskip}{5pt}
\setlength{\belowdisplayskip}{0pt}
\setlength{\abovedisplayshortskip}{0pt}
\setlength{\belowdisplayshortskip}{0pt}
\begin{align}
\hspace*{-0.15in} \begin{array}{ll}
\displaystyle \minimize_{\mb{C}, \mb{B}} & -\log\det\paren{\mb{C} -\mb{B}\widehat{\mb{\Theta}}_{2}\mb{B}^{T}} +   \alpha \norm{\mb{C}}{1}  \\
     & + \text{tr}\paren{\widehat{\mb{\Sigma}}_{1}\paren{ \mb{C} - \mb{B}\widehat{\mb{\Theta}}_{2}\mb{B}^{T}}}  + \beta \norm{ \widehat{\mb{\Theta}}_2\mb{B}^{T}}{2,1} \\
     \st & \mb{C} - \mb{B}\widehat{\mb{\Theta}}_{2}\mb{B}^{T} \succeq \mb{0},
\end{array}
\hspace*{-0.1in}\label{eqn: latent_inv_cov_estimate_nonconvex}
\end{align}
\end{normalsize}
\hspace*{-0.052in}where $\alpha$ and  $\beta$ are positive regularization parameters, 
$\widehat{\mb{\Theta}}_2 \succeq 0$ is the available summary information about $\mb{\Theta}_2$
and $\small \norm{\widehat{\mb{\Theta}}_2\mb{B}^{T}}{2,1}:= \sum_{i}\|(\widehat{\mb{\Theta}}_2\mb{B}^{T})_{i}\|_{2}$ is the  $\ell_{2,1}$ norm that  induces row sparsity of  $\widehat{\mb{\Theta}}_{21}$. 

{The {differences between LV-GGM and} the proposed DiLat-GGM are three-fold. First, compared to the blind LV-GGM,  the semiblind DiLat-GGM utilizes the  information about external network structure  and takes into account their influence on the target network. Second, DiLat-GGM explicitly learns the linear mapping $\mb{B}$, which enables us to  estimate the hidden variables via the conditional mean   $\mb{\mu}_{2|1}$. Thus, it can be used to recover graph signals on $\cV_2$ under GGM. Third, instead of inducing low rank (via nuclear norm) in  \eqref{eq: LV}, a different row-sparsity promoting strategy  is imposed through $\ell_{2,1}$ norm, which explicitly drops irrelevant features during the training. 
{However, unlike LV-GGM,  DiLat-GGM does not lead to a convex optimization problem}. In the next section, we propose an efficient optimization method to solve problem \eqref{eqn: latent_inv_cov_estimate_nonconvex}}. \vspace{-20pt}


\section{Optimization Method}\label{sec: algo_chap3}\vspace{-15pt}
We begin by reformulating   problem \eqref{eqn: latent_inv_cov_estimate_nonconvex} as below 
\begin{normalsize}\setlength{\abovedisplayskip}{5pt}
\setlength{\belowdisplayskip}{0pt}
\setlength{\abovedisplayshortskip}{0pt}
\setlength{\belowdisplayshortskip}{0pt}
\begin{align}
 \hspace*{-0.15in} \begin{array}{ll}
\displaystyle \minimize_{\mb{C}, \mb{B}} & -\log\det \mb{R}(\mb{C}, \mb{B}; \widehat{\mb{\Theta}}_{2})  +   \alpha \norm{\mb{C}}{1}  \\
     & + \text{tr}\paren{\widehat{\mb{\Sigma}}_{1}\paren{ \mb{C} - \mb{B}\widehat{\mb{\Theta}}_{2}\mb{B}^{T}}}  + \beta \norm{ \widehat{\mb{\Theta}}_2\mb{B}^{T}}{2,1} \\
     \st &  \mb{R}(\mb{C}, \mb{B}; \widehat{\mb{\Theta}}_{2})  \succeq \mb{0},
\end{array}
 \hspace*{-0.2in} \label{eqn: latent_inv_cov_estimate_nonconvex_v2}
\end{align}
\end{normalsize} \hspace*{-0.052in}where 
$\small \mb{R}(\mb{C}, \mb{B}; \widehat{\mb{\Theta}}_{2}) :=\small \brac{\begin{smallmatrix}
 \mb{C} & \mb{B} \\ 
\mb{B}^{T} &\widehat{\mb{\Theta}}_{2}^{-1}
\end{smallmatrix} }$ is linear in $(\mb{C}, \mb{B})$, and we have used the fact that
$\small \mb{R}(\mb{C}, \mb{B}; \widehat{\mb{\Theta}}_{2})  \succeq \mb{0}$ is equivalent to $\small  \mb{C} - \mb{B}\widehat{\mb{\Theta}}_{2}\mb{B}^{T} \succeq \mb{0}$ under $\small  \widehat{\mb{\Theta}}_{2} \succeq 0$, as provided by the Shur complement theorem \cite{boyd2004convex}.
Note that 
problem \eqref{eqn: latent_inv_cov_estimate_nonconvex_v2} is non-convex, since $\small \text{tr}\paren{\widehat{\mb{\Sigma}}_{1}\paren{ \mb{C} - \mb{B}\widehat{\mb{\Theta}}_{2}\mb{B}^{T}}}$ is a difference of convex (DC) functions $\small \text{tr}(\hat{\boldsymbol \Sigma}_1 \mathbf C)$  and $\small g(\mb{B}):= \text{tr}(\hat{\boldsymbol \Sigma}_1 \mb{B}\widehat{\mb{\Theta}}_{2}\mb{B}^{T} )$, where the former is the linear with respect to  $\small \mathbf C$, and the latter is quadratic  with respect to   $\small \mathbf B$.

%


Due to the DC-type nonconvexity,  \eqref{eqn: latent_inv_cov_estimate_nonconvex_v2} can be solved  using the convex-concave procedure (CCP) \cite{lipp2016variations, yuille2002concave}. 
Specifically, at each iteration of CCP, it convexifies the concave function $-g(\mb{B})$ through linearization  
\begin{align}
\small \tilde{g}(\mb{B}; \mb{B}_t)= g(\mb{B}_t) + \tr{\nabla_{\mb{B}}g(\mb{B}_t)^{T}\paren{\mb{B}- \mb{B}_t}},
\label{eq: linear}
\end{align} 
where $\small g(\mb{B}) = \text{tr}\paren{\widehat{\mb{\Sigma}}_{1}\mb{B}\widehat{\mb{\Theta}}_{2}\mb{B}^{T}} $ in \eqref{eqn: latent_inv_cov_estimate_nonconvex_v2}, $t$ is the iteration index of CCP, and
$\small \nabla_{\mb{B}}g(\mb{B}_t) =  2\widehat{\mb{\Sigma}}_{1}\mb{B}_{t}\widehat{\mb{\Theta}}_{2}$
yields the gradient of $\small g(\mb{B})$ at point $\mathbf B_t$. Upon defining $\small \mb{D}_t:= \mb{B}_{t}\widehat{\mb{\Theta}}_2$ and substituting  \eqref{eq: linear} into \eqref{eqn: latent_inv_cov_estimate_nonconvex_v2}, 
CCP iteratively solves the {convex program}, 
\begin{normalsize}\setlength{\abovedisplayskip}{5pt}
\setlength{\belowdisplayskip}{4pt}
\setlength{\abovedisplayshortskip}{0pt}
\setlength{\belowdisplayshortskip}{0pt}
\begin{align}
\hspace*{-0.2in}\begin{array}{ll}
    \displaystyle \minimize_{\mathbf C, \mathbf B} & -\log\det\mb{R}(\mb{C}, \mb{B}; \widehat{\mb{\Theta}}_{2}) + \text{tr}\paren{\widehat{\mb{\Sigma}}_{1}\paren{\mb{C} - 2\mb{B}\mb{D}_{t}^{T}}}  \\
     & +\alpha_{m} \norm{\mb{C}}{1} + \beta_{m} \norm{ \widehat{\mb{\Theta}}_2\mb{B}^{T}}{2,1} \\
     \st & \mb{R}(\mb{C}, \mb{B}; \widehat{\mb{\Theta}}_{2})  \succeq \mb{0}.
\end{array}\hspace*{-0.5in}
\label{eqn: latent_inv_cov_estimate_nonconvex_sub}
\end{align}
\end{normalsize}  \hspace*{-0.08in} Problem \eqref{eqn: latent_inv_cov_estimate_nonconvex_sub} can be solved using a semidefinite programming solver, e.g., provided by CVX \cite{grantcvx}. However, 
this leads to high computational complexity  $O(n^{6.5})$.
To improve computation efficiency, the alternating direction method of multipliers (ADMM)  \cite{boyd2011distributed}  will be used to solve problem \eqref{eqn: latent_inv_cov_estimate_nonconvex_sub}.
By introducing  auxiliary variables $\small \mb{R}:=\mb{R}(\mb{C}, \mb{B}; \widehat{\mb{\Theta}}_{2}) ,  \mb{P}:= \brac{\begin{array}{cc}
\mb{P}_1 & \mb{P}_{21}^{T} \\ 
\mb{P}_{21} & \mb{P}_{2}
\end{array} } = \mb{R}$ and $\small \mb{W} := \widehat{\mb{\Theta}}_2\mb{P}_{21}$,  we  rewrite problem \eqref{eqn: latent_inv_cov_estimate_nonconvex_sub} as 
\begin{normalsize}\setlength{\abovedisplayskip}{2pt}
\setlength{\belowdisplayskip}{0pt}
\setlength{\abovedisplayshortskip}{0pt}
\setlength{\belowdisplayshortskip}{0pt}
\begin{align}
\hspace*{-0.4in}\begin{array}{ll}
    \displaystyle \minimize_{\mb{R}, \mb{P},\mb{W}}& -\log\det\mb{R} + \text{tr}\paren{\mb{S}\mb{R}} +\alpha \norm{\mb{P}_1}{1}     \\
    &+ \beta \norm{\mb{W}}{2,1}+ \ind{\mb{R} \succeq \mb{0}}\\
\st& \mb{R} = \mb{P}, \mb{P}_{2} =\widehat{\mb{\Theta}}_{2}^{-1}, \mb{W} = \widehat{\mb{\Theta}}_2\mb{P}_{21}, 
\end{array}\hspace*{-0.4in}
\label{eqn: admm}
\end{align}\end{normalsize} \hspace*{-0.12in}   where  $\small \mb{S}:=  \brac{\begin{smallmatrix}
\widehat{\mb{\Sigma}}_{1} & -\widehat{\mb{\Sigma}}_{1}\mb{D}_{t} \\ 
-\mb{D}_{t}^{T}\widehat{\mb{\Sigma}}_{1}   & \gamma_{t} \mb{I}
\end{smallmatrix} },$
$\small \ind{A}$ is an indicator function of set $A$. {Problem \eqref{eqn: admm} now fits  the standard form of ADMM, a convex program with equality constraints.} Based on the augmented Lagrangian \eqref{eqn: admm}, ADMM leads to
three subproblems with respect to $\mb{R}, \mb{P}$ and $\mb{W}$, respectively.  The proposed ADMM-based algorithm yields the   complexity $O(n^{3})$ due to the SVD step while solving the subproblem with respect $\mb{R}$  \cite{boyd2011distributed}. It is also shown in  \cite{lanckriet2009convergence} that CCP converges to a local stationary point regardless of choice of initial points. Here we choose the initial point $(\mb{C}_0, \mb{B}_0)$  heuristically according to  \cite{xu2017speeding}. 
We summarize the CCP-based algorithm to solve DiLat-GGM    in \textbf{Algorithm \ref{alg: dilat_ggm_ccp}} \vspace{-15pt}

\begin{algorithm}[t]
  \caption{DiLat-GGM via Convex-concave procedure}
  \label{alg: dilat_ggm_ccp}
  \begin{algorithmic}[1]
  \small{ 
  \REQUIRE Marginal covariance $\widehat{\mb{\Sigma}}_{1}$ of $\mb{x}_1 \in \bR^{n_1}$. The parameters  $\alpha, \beta >0$. A noisy summary matrix $\widehat{\mb{\Theta}}_2  \succ \mb{0}\in \bR^{n_2\times n_2}$ of $\mb{x}_2$.  

  \STATE \textbf{Initialize:} Random initialization or by heuristic according to \cite{xu2017speeding}. Return $(\mb{C}_{0}, \mb{B}_{0})$.
  
  \FOR{$t=1,\ldots, T$ or until converge} \vspace{1pt} 
    \STATE Construct matrix 
$\small \mb{D}_{t-1} := \mb{B}_{t-1}\widehat{\mb{\Theta}}_2$ 

    \STATE Solve the  subproblem \eqref{eqn: admm} via ADMM. Return $\footnotesize (\mb{C}_{t}, \mb{B}_{t})$.
\ENDFOR  %
\vspace{2pt}
\ENSURE 
   Output $\small \mb{C}_{T}:= [\mb{R}_{T}]_{\cV_1\times \cV_1}$ and $\mb{B}_{T}:= [\mb{R}_{T}]_{\cV_1\times \cV_2}$. 
 }
 \end{algorithmic} 
\end{algorithm}
\section{Experiments}\label{sec: experiments_chap3}\vspace{-15pt}
In this section, we compare the performance of the semiblind \textbf{DiLat-GGM} 
with three blind graph topology learning algorithms: the graphical Lasso (\textbf{GLasso}) \cite{friedman2008sparse};  the latent variable  Gaussian graphical model (\textbf{LV-GGM}) \cite{chandrasekaran2012latent} and the generalized Laplacian learning (\textbf{GLap}) \cite{pavez2016generalized} which is a variant of GLasso. 

\setlength{\belowcaptionskip}{-5pt}
\begin{figure*}[tb]
\centering
  \begin{minipage}{0.33\linewidth}
  \centering
    \begin{minipage}[b]{0.15\linewidth}
  \centering
  \centerline{\includegraphics[scale = 0.17]{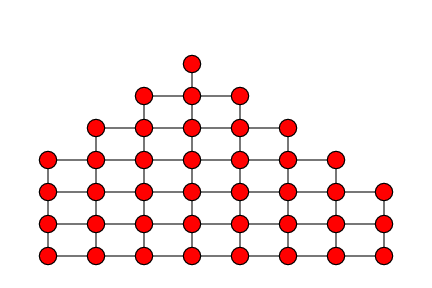}}
  \vspace{-3pt}
  \centerline{\scriptsize (a) Ground truth}
  \end{minipage}\hspace{0.35\linewidth}
  \begin{minipage}[b]{0.15\linewidth}
  \centering
  \centerline{\includegraphics[scale = 0.17]{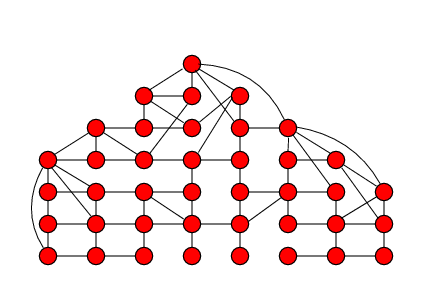}}
    \vspace{-3pt}
  \centerline{\scriptsize (b) GLasso}
  \end{minipage}\\
   \begin{minipage}[b]{0.15\linewidth}
  \centering
  \centerline{\includegraphics[scale = 0.17]{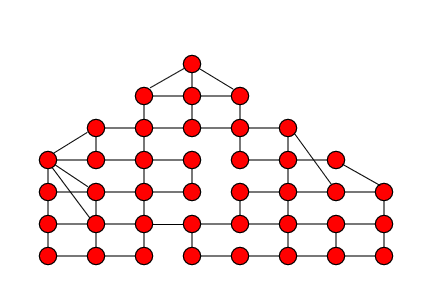}}
  \vspace{-5pt}
  \centerline{\scriptsize (c) LV-GGM}
  \end{minipage}\hspace{0.35\linewidth}
  \begin{minipage}[b]{0.15\linewidth}
  \centering
  \centerline{\includegraphics[scale = 0.17]{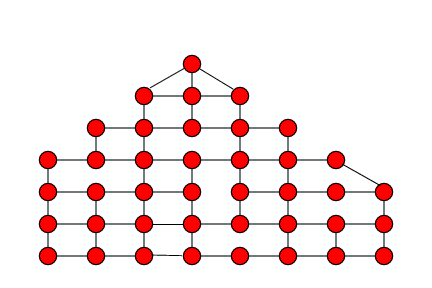}}
    \vspace{-5pt}
  \centerline{\scriptsize (d) \textbf{DiLat-GGM}}
  \end{minipage}
  \end{minipage}
 \begin{minipage}{0.32\linewidth}
  \centering
  \centerline{\includegraphics[scale=0.33]{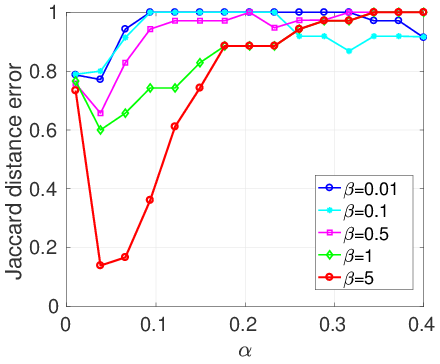}}
  \vspace{-8pt}
  \centerline{(e)}
  \end{minipage}
   \begin{minipage}{0.33\linewidth}
  \centering
  \centerline{\includegraphics[scale=0.32]{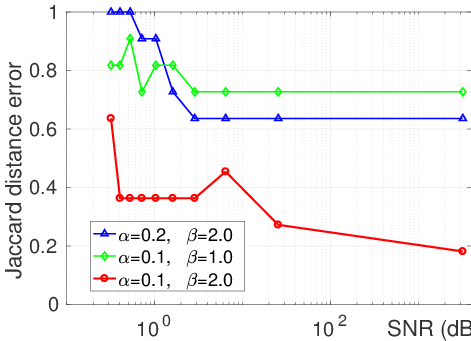}}
  \vspace{-8pt}
  \centerline{(f)}
  \end{minipage}
  \caption{\scriptsize (a)-(d) A comparison of sub-network topology. (a) The ground truth of size $n_1=40$ with grid structure {where the unobserved latent variables (not shown) have similar grid dependencies.} (b) The graph learned by \textbf{GLasso} with optimal $\alpha=0.4$ 
  (c) The graph learned by \textbf{LV-GGM} with optimal $\alpha=0.1, \beta=0.15$  (d) The graph learned by \textbf{DiLat-GGM} with optimal $\alpha=0.2, \beta=1$. GLasso has high false positives (cross-edges between leaves) due to the marginalization effect. Compare to LV-GGM, the DiLat-GGM has fewer missing edges and less false positives.  (e) The $\alpha-$sensitivity of DiLat-GGM under the grid network model in (a) for  different choices of regularization parameters. The performance is measured in terms of Jaccard distance error. (f) The robustness of DiLat-GGM for grid network (a) under different $(\alpha, \beta)$ when $\widehat{\mb{\Theta}}_2=  \widehat{\mb{L}}_2 + \sigma_{L}^2 \mb{G}$ with the Signal-to-Noise Ratio (SNR) $\|\widehat{\mb{L}}_2\|_{F}^2/\sigma_{L}^2$.}\label{fig: alpha_beta_theta}
\end{figure*}

In the following experiments, {we generate a network $\cG= (\cV, \cE)$ with $|\cV|=n$ and then compute the normalized Laplacian matrix $\mb{L}$. } The random graph signal $\mb{x}\in \bR^{n}$ is drawn from $\cN(\mb{0},  (\mb{L} + \epsilon\mb{I})^{-1})$ where $\epsilon=10^{-3}$. We compare over different graph topologies, including the {complete binary tree} with {height} $h$, the {grid network} with {width} $w$ and {height} $h$ and the  {Erd\H{o}s-R\'{e}nyi graph} with {size} $n$ and {edge probability} $p$. The vertex set $\cV_1$ of sub-network $\cG_{1}$ is sampled randomly from $\cV$ with $|\cV_{1}|=n_1$. The edge set $\cE_1= \cE \cap (\cV_1 \times \cV_1)$.  {The data sample matrix $\mb{X}$ is composed of $m$ i.i.d realizations of $\mb{x}$, where we choose $m = 500$.} 

\begin{table}[tb]
\caption{\footnotesize Topology estimation error for different graphs, with the best performance shown in \textbf{bold}.} \vspace{-9pt}
\label{tab: edge_accuracy}
 \centering 
 \footnotesize
\begin{tabular}{|m{95pt}<{\centering}|m{22pt}<{\centering}|m{18pt}<{\centering}|m{18pt}<{\centering}|m{18pt}<{\centering}|}
\hline 
\multicolumn{5}{|c|}{Mean Jaccard distance error ($\times 100\%$)  } \\ 
\hline
 Network &  GLasso  &  GLap &   LV-GGM &  DiLat-GGM \\ 
\hline 
\textbf{complete binary tree} ($h=3, n_1=10$)  & $55.7$ & $\mb{12.8}$  &  $36.4$   &  $18.8 $  \\ 
\hline 
\textbf{complete binary tree} ($h=5, n_1=36$)  & $15.0$ & $50.9$   &  $3.3 $   &  $\mb{2.5}$  \\ 
\hline 
\textbf{grid} ($w=5, h=5, n_1=15$) & $39.3$ & $\mb{5.7} $ &  $23.3 $   & $12.8$  \\ 
\hline 
\textbf{grid} ($w=9, h=9, n_1=49$) & $10.3$ & $32.7  $  &  $7.8 $    & $\mb{5.4}$\\ 
\hline 
\textbf{Erd\H{o}s-R\'{e}nyi} ($n=15, p=0.05, n_1=10$)  & $19.6$ &  $\mb{7.9}$  & $15.0$ & $13.9$\\ 
\hline 
\textbf{Erd\H{o}s-R\'{e}nyi} ($n=60, p=0.05, n_1=40$) & $10.8$ & $61.1$  &  $8.1$ &$\mb{6.5}$\\
\hline 
\end{tabular} \vspace{-1pt}
\end{table}

{To measure the accuracy of semiblind to blind topology inference (as compared with ground truth), we consider }
the {Jaccard distance} \cite{choi2010survey} between two sets $A, B$ $
\small dist_{J}(A, B) = 1 - \frac{\abs{A\cap B}}{\abs{A\cup B}} \in [0,1]
$, where $A$ is the support set of the estimated sparse matrix $\small \widehat{\mb{C}}$, and $B$ {is given by the edge set of the ground-truth $\cG_1$.}
The Jaccard distance is a widely used similarity measure in information retrieval \cite{manning2008introduction}.  
For DiLat-GGM, we choose the matrix $\small \widehat{\mb{\Theta}}_2$ as $\small \widehat{\mb{\Theta}}_2=  \widehat{\mb{L}}_2 + \sigma_{L}^2 \mb{G},$ where $\small \widehat{\mb{L}}_2$  is an estimate of precision matrix over $\mb{x}_2$, $\sigma_{L}\in [0.01, 10]$ and $\small \mb{G} =\mb{H}\mb{H}^{T}/(n_2)^2$ is a Gram matrix generated by Gaussian random matrix $\small \mb{H} \in \bR^{n_2 \times n_2}$ with $\small \mb{H}_{i,j} \stackrel{i.i.d.}{\sim} \cN(0,1)$. {The noise term $\sigma_{L}^2\mb{G}$ can be interpreted as a noise introduced by finite sample errors in the sample precision matrix estimator $\widehat{\mb{L}}_2$ or it may be intentionally introduced to increase privacy.}

Table \ref{tab: edge_accuracy} shows the topology estimation error under different graphs for GLasso, GenLap, LV-GGM and DiLat-GGM in terms of the Jaccard distance. All results are based on an average of $50$ runs and for each run we choose the best performance after a grid search of regularization parameter $\alpha \in [10^{-2}, 0.7]$, $\beta\in [0.01, 5], \sigma\in [0.01, 5]$. 
As we can see, our proposed semiblind DiLat-GGM reaches superior performance compared to the blind GLasso, LV-GGM for all investigated networks. For small network, a worse performance of DiLat-GGM is  due to a less accurate solution of CCP for solving non-convex problems. However, in large network, DiLat-GGM exploits more hidden structures in topology estimation, and thus yields a better performance. 

{In Figure \ref{fig: alpha_beta_theta}  (a)-(d), we show the learned network resulting from GLasso, LV-GGM and DiLat-GGM under different choices of optimal parameters $\alpha, \beta$.}
As we can see, {DiLat-GGM edge estimates} have lower miss rate and false positive rate  compared to GLasso and LV-GGM. The GLasso, however, has a higher false positive rate in boundary vertices due to the effect of marginalization bias.

In Figure \ref{fig: alpha_beta_theta} (e), we demonstrate the sensitivity of the DiLat-GGM model under different choices of regularization parameter $\alpha$ and $\beta$. The underlying network is shown in Figure \ref{fig: alpha_beta_theta} (a). For DiLat-GGM, $\widehat{\mb{\Theta}}_{2} $ is the same as above. {We observe that } 
when $\alpha$ increases, the learned graph becomes overly sparse and Jaccard distance error increases. The choice of $\beta$ controls the row sparsity of the conditional cross precision $\mb{\Theta}_{21}$, if it is too small, the DiLat-GGM cannot capture the local effect of the latent variables, which decreases its performance in sub-network learning.  
In Figure \ref{fig: alpha_beta_theta} (f), we evaluate the robustness of DiLat-GGM when the pre-defined matrix $\widehat{\mb{\Theta}}_2$ is corrupted by {noise of different levels}. As expected, when the Signal-to-Noise Ratio (SNR) $\|\widehat{\mb{L}}_2\|_{F}^2/\sigma_{L}^2$ decreases, the performance of DiLat-GGM decreases. However, DiLat-GGM is robust for a wide range of choice of $\sigma_{L}$. 
\vspace{-18pt}


\section{Conclusion}\label{sec: conclusion_chap3}\vspace{-20pt}
In this paper, we proposed {a semiblind subgraph estimation algorithm called} DiLat-GGM that learns a sparse sub-network topology with {noisy information about the network topology external to the subgraph}. We show that DiLat-GGM leads to a DC-type nonconvex optimization problem, whose local optimal solution was computed via an efficient CCP-based algorithm. {Extensive numerical results show that the proposed semiblind DiLat-GGM outperforms  the state-of-the-art blind sparse GGMs in terms of topology estimation accuracy}.
In the future, we will apply the proposed algorithm to  larger scale datasets and {develop a strategy to estimate both $\cG_1$ and $\cG_2$}. 

\bibliographystyle{IEEEbib}
\bibliography{xie_globalsip17.bib}

\newpage
\section{Appendix}
\subsection{Solving subproblem \eqref{eqn: latent_inv_cov_estimate_nonconvex_sub} using ADMM}\label{sec: admm}
Following the ADMM procedure, we form an augmented Lagrangian for problem as
\begin{small} \setlength{\abovedisplayskip}{5pt}
\setlength{\belowdisplayskip}{0pt}
\setlength{\abovedisplayshortskip}{0pt}
\setlength{\belowdisplayshortskip}{0pt}
\begin{align*}
\begin{array}{ll}
  \displaystyle
&\cL(\mb{R}, \mb{P})\\ 
&= -\log\det\mb{R} + \text{tr}\paren{\mb{S}\mb{R}}  +\alpha \norm{\mb{P}_1}{1} + \beta \norm{\mb{W}}{2,1}+ \ind{\mb{R} \succeq \mb{0}}\\
& + \ind{\mb{P}_{2} =\widehat{\mb{\Theta}}_{2}^{-1}}+  \tr{\mb{\Lambda}^{T}(\mb{R} - \mb{P})} + \frac{\rho}{2}\norm{\mb{R} - \mb{P}}{F}^{2}\\
& + \tr{\mb{\Lambda}_w^{T}\paren{\mb{W} - \widehat{\mb{\Theta}}_{2}\mb{P}_{21}}} + \frac{\rho_w}{2}\norm{\mb{W} - \widehat{\mb{\Theta}}_{2}\mb{P}_{21}}{F}^2,
\end{array}
\end{align*}
\end{small}\hspace*{-0.052in} where $\mb{\Lambda} \in \bR^{n\times n}$ and $\mb{\Lambda}_{w}$ form dual matrices. ADMM minimizes the augmented Lagrangian via block coordinate descent. In specific, it solves two separable problems: 
\begin{small} \setlength{\abovedisplayskip}{5pt}
\setlength{\belowdisplayskip}{0pt}
\setlength{\abovedisplayshortskip}{0pt}
\setlength{\belowdisplayshortskip}{0pt}
 \begin{align}
 \hspace*{-0.04in} \begin{array}{ll}
  \displaystyle \minimize_{\mb{R}}& -\log\det\mb{R} + \text{tr}\paren{\mb{S}\mb{R}}   \\
  &+ \tr{\mb{\Lambda}^{T}(\mb{R} - \mb{P})} + \frac{\rho}{2}\norm{\mb{R} - \mb{P}}{F}^{2}  + \ind{\mb{R} \succeq \mb{0}}\\
&= -\log\det\mb{R} + \text{tr}\paren{\mb{S}\mb{R}} +    \frac{\rho}{2}\norm{\mb{R} - \mb{P} + \frac{1}{\rho}\mb{\Lambda}}{F}^{2}\\
&+ \ind{\mb{R} \succeq \mb{0}}, 
\end{array} 
\hspace*{-0.04in} \label{eqn: latent_inv_cov_estimate_nonconvex_subsub1}
\end{align}\end{small}
and 
\begin{small} \setlength{\abovedisplayskip}{5pt}
\setlength{\belowdisplayskip}{0pt}
\setlength{\abovedisplayshortskip}{0pt}
\setlength{\belowdisplayshortskip}{0pt}
 \begin{align}
 \begin{array}{ll}
  \displaystyle \minimize_{\mb{P}, \mb{W}}& \alpha \norm{\mb{P}_1}{1}    + \tr{\mb{\Lambda}^{T}(\mb{R} - \mb{P})} + \frac{\rho}{2}\norm{\mb{R} - \mb{P}}{F}^{2}\\
&  + \beta \norm{\mb{W}}{2,1} + \tr{\mb{\Lambda}_w^{T}\paren{\mb{W} - \widehat{\mb{\Theta}}_{2}\mb{P}_{21}}}\\
& + \frac{\rho_w}{2}\norm{\mb{W} - \widehat{\mb{\Theta}}_{2}\mb{P}_{21}}{F}^2+  \ind{\mb{P}_{2} =\widehat{\mb{\Theta}}_{2}^{-1}}\\
&=\alpha \norm{\mb{P}_1}{1}   + \beta \norm{\mb{W}}{2,1} +    \frac{\rho}{2}\norm{ \mb{P} - \mb{R} - \frac{1}{\rho}\mb{\Lambda}}{F}^{2}\\
&+ \frac{\rho_{w}}{2}\norm{ \mb{W} -\widehat{\mb{\Theta}}_{2}\mb{P}_{21} - \frac{1}{\rho_w}\mb{\Lambda}_w}{F}^{2} \\
&+  \ind{\mb{P}_{2} =\widehat{\mb{\Theta}}_{2}^{-1}}. 
\end{array} 
\hspace*{-0.04in} \label{eqn: latent_inv_cov_estimate_nonconvex_subsub2}
\end{align}\end{small}

From Section \ref{sec: algo_chap3}, we see that \eqref{eqn: latent_inv_cov_estimate_nonconvex_subsub1} corresponds to a proximal operator
\begin{small} \setlength{\abovedisplayskip}{5pt}
\setlength{\belowdisplayskip}{0pt}
\setlength{\abovedisplayshortskip}{0pt}
\setlength{\belowdisplayshortskip}{0pt}
\begin{align}
\text{Prox}_{\mb{R}}(\mb{Z}, \xi):= \minimize_{\mb{R} \succ \mb{0}} \frac{1}{2\xi}\norm{\mb{R}- \mb{Z}}{F}^2 -\log\det\left(\mathbf{R}\right)+ \text{tr}\left(\mb{S}\mathbf{R}\right).\label{eqn: latent_ggm_nonconvex_sub_proxR}
\end{align}\end{small}
The optimal solution of above satisfies that the gradient of the objective function
\begin{small} \setlength{\abovedisplayskip}{5pt}
\setlength{\belowdisplayskip}{0pt}
\setlength{\abovedisplayshortskip}{0pt}
\setlength{\belowdisplayshortskip}{0pt}
\begin{align*} 
\frac{1}{\xi}\paren{\mb{R}- \mb{Z}} - \mb{R}^{-1} + \mb{S} &= 0. 
\end{align*}\end{small} Let the eigen-decomposition of $\xi\mb{S} - \mb{Z}:=\mb{U}\diag{\mb{\sigma}}\mb{U}^{T}$, where $\mb{\sigma} := (\sigma_{i})$. Then the optimal solution 
\begin{small} \setlength{\abovedisplayskip}{5pt}
\setlength{\belowdisplayskip}{0pt}
\setlength{\abovedisplayshortskip}{0pt}
\setlength{\belowdisplayshortskip}{0pt}
\begin{align*}
\mb{R}&= \mb{U}\diag{\mb{\gamma}}\mb{U}^{T}\\
\text{where }\gamma_{i}&= \frac{-\sigma_{i} + \sqrt{\sigma_{i}^2 + 4\,\xi} }{2} >0.
\end{align*}\end{small}

To solve \eqref{eqn: latent_inv_cov_estimate_nonconvex_subsub2}, we see that the objective of \eqref{eqn: latent_inv_cov_estimate_nonconvex_subsub2} is separable as well.  Problem \eqref{eqn: latent_inv_cov_estimate_nonconvex_subsub2} is equivalent to
\begin{small} \setlength{\abovedisplayskip}{5pt}
\setlength{\belowdisplayskip}{0pt}
\setlength{\abovedisplayshortskip}{0pt}
\setlength{\belowdisplayshortskip}{0pt}
\begin{align}
\hspace*{-0.04in}  \begin{array}{ll}
  \displaystyle \minimize_{\mb{P}_1, \mb{P}_{21}, \mb{W}}& \phantom{=} \alpha_{m} \norm{\mb{P}_{1}}{1}    +  \frac{\rho}{2}\norm{ \mb{P}_1 - \mb{R}_1 - \frac{1}{\rho}\mb{\Lambda}_{1}}{F}^{2}  \\
&+ \beta_{m} \norm{\mb{W}}{2,1}  + \frac{\rho}{2}\norm{ \mb{P}_{21} - \mb{R}_{21} - \frac{1}{\rho}\mb{\Lambda}_{21}}{F}^{2}\\ 
& +  \frac{\rho_{w}}{2}\norm{ \mb{W} -\widehat{\mb{\Theta}}_{2}\mb{P}_{21} - \frac{1}{\rho_w}\mb{\Lambda}_w}{F}^{2} 
\end{array}
\hspace*{-0.04in} \label{eqn: latent_ggm_nonconvex_sub2_problem}
\end{align}\end{small} and $\mb{P}_2 = \mb{T}$. It involves three proximal operators: first, 
\begin{small} \setlength{\abovedisplayskip}{5pt}
\setlength{\belowdisplayskip}{0pt}
\setlength{\abovedisplayshortskip}{0pt}
\setlength{\belowdisplayshortskip}{0pt}
\begin{align*}
\text{Prox}_{\mb{P}_{1}, \alpha}(\mb{Z}, \xi):= \minimize_{\mb{P}_{1}} \frac{1}{2\xi}\norm{\mb{P}_{1}- \mb{Z}}{F}^2 + \alpha \norm{\mb{P}_{1}}{1}
\end{align*}\end{small}
which is equivalent to 
\begin{small} \setlength{\abovedisplayskip}{5pt}
\setlength{\belowdisplayskip}{0pt}
\setlength{\abovedisplayshortskip}{0pt}
\setlength{\belowdisplayshortskip}{0pt}
\begin{align}
\text{Prox}_{\mb{P}_{1}, \alpha}(\mb{Z}, \xi)&= \text{soft-threshold}(\mb{Z}, \xi\,\alpha). \label{eqn: latent_ggm_nonconvex_sub_proxC}
\end{align}\end{small} Second,
\begin{small} \setlength{\abovedisplayskip}{5pt}
\setlength{\belowdisplayskip}{0pt}
\setlength{\abovedisplayshortskip}{0pt}
\setlength{\belowdisplayshortskip}{0pt}
\begin{align*}
\hspace*{-0.04in}  \begin{array}{ll}
  \displaystyle 
&\text{Prox}_{\mb{P}_{21}}(\mb{Z}, \mb{Z}^{'}, \xi, \xi_{w})\\
&:= \minimize_{\mb{P}_{21}} \frac{1}{2\xi}\norm{ \mb{P}_{21} -\mb{Z}}{F}^{2} + \frac{1}{2\xi_{w}}\norm{\widehat{\mb{\Theta}}_2\mb{P}_{21}- \mb{Z}^{'}}{F}^{2}
\end{array}
\end{align*}\end{small} This is a linear transformation
\begin{small} \setlength{\abovedisplayskip}{5pt}
\setlength{\belowdisplayskip}{0pt}
\setlength{\abovedisplayshortskip}{0pt}
\setlength{\belowdisplayshortskip}{0pt} 
\begin{align}
&\phantom{=}\text{Prox}_{\mb{P}_{21}}(\mb{Z}, \mb{Z}^{'}, \xi, \xi_{w})\nonumber\\
&= \paren{\xi_{w}\mb{I}+\xi\widehat{\mb{\Theta}}_2^2}^{-1}\paren{\xi_{w}\mb{Z} +  \xi\widehat{\mb{\Theta}}_2\mb{Z}^{'}} \nonumber\\
&= \mb{U}\text{diag}\brac{\frac{\xi_{w}}{\xi_{w}+ \xi \lambda_{i}^{2}}}_{i,i}\mb{U}^{T}\mb{Z}+ \mb{U}\text{diag}\brac{\frac{ \xi \lambda_i}{\xi_{w}+ \xi \lambda_{i}^{2}}}_{i,i}\mb{U}^{T}\mb{Z}^{'}  \label{eqn: latent_ggm_nonconvex_sub_proxB_c2}
\end{align}\end{small}\hspace*{-0.04in}   where  $\widehat{\mb{\Theta}}_2= \mb{U}\text{diag}\brac{\lambda_i}_{i,i}\mb{U}^{T}$ is the eigen-decomposition.  
And the proximal operator
\begin{small} \setlength{\abovedisplayskip}{5pt}
\setlength{\belowdisplayskip}{0pt}
\setlength{\abovedisplayshortskip}{0pt}
\setlength{\belowdisplayshortskip}{0pt}  
\begin{align*}
\text{Prox}_{\mb{W}, \beta}(\mb{Z}^{'}, \xi):=\minimize_{\mb{W}\in \bR^{n_2\times n_1}}&\phantom{=} \frac{1}{2\xi}\norm{\mb{W} -\mb{Z}^{'}}{F}^{2} + \beta \norm{\mb{W}}{2,1},
\end{align*}\end{small}\hspace*{-0.02in} which has optimal solution $\mb{W}$  with $i$-th row 
\begin{small} \setlength{\abovedisplayskip}{5pt}
\setlength{\belowdisplayskip}{0pt}
\setlength{\abovedisplayshortskip}{0pt}
\setlength{\belowdisplayshortskip}{0pt}  
\begin{align}
\mb{W}_{i}&= \paren{1 - \frac{\beta \xi}{\norm{\mb{Z}^{'}_{i}}{2}}}_{+}\mb{Z}^{'}_{i}, \quad i=1,\ldots, n_2  \label{eqn: latent_ggm_nonconvex_sub_proxW}
\end{align}\end{small}

Finally, we have the dual updates
\begin{align*}
\mb{\Lambda}^{(t)} := \mb{\Lambda}^{(t-1)} + \rho\paren{\mb{R}- \mb{P}}\\
\mb{\Lambda}_{w}^{(t)} := \mb{\Lambda}_{w}^{(t-1)} + \rho_w\paren{\mb{W} -\widehat{\mb{\Theta}}_{2}\mb{P}_{21}}
\end{align*}
The algorithm of ADMM is summarized in Algorithm \ref{alg: dilat_ggm_admm}.

\begin{algorithm*}[t]
  \caption{DiLat-GGM subproblem  via ADMM}
  \label{alg: dilat_ggm_admm}
  \begin{algorithmic}[1]
  \small{ 
  \REQUIRE Positive definite matrix  $\mb{S} \succ \mb{0}$ and $\mb{S} \in \bR^{n\times n}$. The nonnegative regularization parameter  $\alpha, \beta >0$. The pre-defined nonegative definite matrix $\widehat{\mb{\Theta}}_2  \succeq \mb{0}$ and $\widehat{\mb{\Theta}}_2\in \bR^{n_2\times n_2}$. Let $\mb{T} = \widehat{\mb{\Theta}}_2^{-1}.$  Let $n_1 = n-n_2$. Dual update parameter $\mu, \mu_{w} >0$. \vspace{5pt}

  \STATE \textbf{Initialize:} Choose an random matrix $\mb{R}^{(0)}=  \brac{\begin{array}{cc}
  \mb{R}^{(0)}_{1} & \mb{R}^{(0)}_{12} \\ 
  \mb{R}^{(0)}_{21} & \mb{R}^{(0)}_{2}
  \end{array} } \in \bR^{n\times n}$ and $\mb{R}^{(0)}\succ \mb{0}$. $\mb{\Lambda}^{(0)} = \mb{0}\in \bR^{n\times n} = \brac{\begin{array}{cc}
  \mb{\Lambda}^{(0)}_{1} & \mb{\Lambda}^{(0)}_{12} \\ 
  \mb{\Lambda}^{(0)}_{21} & \mb{\Lambda}^{(0)}_{2}
  \end{array} }$.  $\mb{\Lambda}^{(0)}_{W} = \mb{0} \in \bR^{n_2\times n_1}$. Let  $\mb{P}^{(0)}  = \brac{\begin{array}{cc}
  \mb{P}^{(0)}_{1} & \mb{P}^{(0)}_{12} \\ 
  \mb{P}^{(0)}_{21} & \mb{P}^{(0)}_{2}
  \end{array} } = \mb{R}\in \bR^{n\times n}.$ Choose $\mb{W}^{(0)} = \widehat{\mb{\Theta}}_{2}\mb{P}_{21}^{(0)}$. \vspace{5pt}
  
  \FOR{$t=1,\ldots, T$ or until converge} \vspace{5pt} 
    \STATE Find  $\mb{P}_1^{(t)} \in  \bR^{n_1\times n_1}$ via $\mb{P}_1^{(t)}= \text{Prox}_{\mb{P}_1, \alpha}(\mb{R}^{(t-1)}_1+\mu\mb{\Lambda}^{(t-1)}_1, \mu)$ as in \eqref{eqn: latent_ggm_nonconvex_sub_proxC}; \vspace{5pt}
    
    \IF{$\widehat{\mb{\Theta}}_2:= \diag{\widehat{\mb{\Theta}}_2}$}
           \STATE Find $\mb{P}_{21}^{(t)}  \in  \bR^{n_2\times n_1}$ via $\mb{P}_{21}^{(t)} = \text{Prox}^{'}_{\mb{P}_{21}, \beta}(\mb{R}^{(t-1)}_{21}+\mu\mb{\Lambda}^{(t-1)}_{21}, \mu)$ as in \eqref{eqn: latent_ggm_nonconvex_sub_proxB_c1} \vspace{3pt}
           
    \ELSE
           \STATE Find  $\mb{W}^{(t)} \in  \bR^{n_2\times n_1}$ via $\mb{W}^{(t)}= \text{Prox}_{\mb{W}, \beta}(\widehat{\mb{\Theta}}_{2}\mb{P}_{21}^{(t-1)}- \mu_{w}\mb{\Lambda}_{W}^{(t-1)}, \mu_{w})$ as in \eqref{eqn: latent_ggm_nonconvex_sub_proxW};   \vspace{5pt}
             
           \STATE Find  $\mb{P}_{21}^{(t)} =\text{Prox}_{\mb{P}_{21}}(\mb{R}^{(t-1)}_{21}+\mu\mb{\Lambda}^{(t-1)}_{21},\mb{W}^{(t)}+\mu_w\mb{\Lambda}_{W}^{(t-1)}, \mu, \mu_{w})$ as  in \eqref{eqn: latent_ggm_nonconvex_sub_proxB_c2};  \vspace{3pt}
           
           \STATE Update dual variables $ \mb{\Lambda}_{W}.$
               \setlength{\abovedisplayskip}{1pt}
                \setlength{\belowdisplayskip}{0.5pt} 
                \begin{align*}
                \mb{\Lambda}_{W}^{(t)} &= \mb{\Lambda}_{W}^{(t-1)} + \frac{1}{\mu_{w}}\paren{\mb{W}^{(t)} -\widehat{\mb{\Theta}}_{2}\mb{P}^{(t)}_{21}}
                \end{align*} 
    \ENDIF \vspace{5pt}
    
    \STATE Set $\mb{P}^{(t)}_2 = \mb{T}$  and $\mb{P}_{12}^{(t)} = \paren{\mb{P}_{21}^{(t)}}^{T}.$ Construct $\mb{P}^{(t)}$.\vspace{5pt}
  
    \STATE Find $\mb{R}^{(t)} \in  \bR^{n\times n}$ via $\mb{R}^{(t)}= \text{Prox}_{\mb{R},\alpha}(\mb{P}^{(t)}- \mu \mb{\Lambda}^{(t-1)}, \mu)$ as in \eqref{eqn: latent_ggm_nonconvex_sub_proxR}. \vspace{5pt}
   
    \STATE Update dual variables $\mb{\Lambda}$
    \setlength{\abovedisplayskip}{1pt}
    \setlength{\belowdisplayskip}{0.5pt} 
    \begin{align*}
    \mb{\Lambda}^{(t)} &= \mb{\Lambda}^{(t-1)} + \frac{1}{\mu}\paren{\mb{R}^{(t)} - \mb{P}^{(t)}}.
    \end{align*}    
\ENDFOR  %
\vspace{2pt}
\ENSURE 
   Output $(\mb{R}^{(T)}, \mb{P}^{(T)})$ if $\widehat{\mb{\Theta}}_2$ is diagonal and $(\mb{R}^{(T)}, \mb{P}^{(T)}, \mb{W}^{(T)})$ otherwise.
 }
 \end{algorithmic} 
\end{algorithm*}
\setlength{\textfloatsep}{1.2em} 
\end{document}